\documentclass[10pt,twocolumn,letterpaper]{article}
\usepackage[accsupp]{axessibility}
\usepackage{cvpr}
\usepackage{times}
\usepackage{epsfig}
\usepackage{graphicx}
\usepackage{svg}
\usepackage{amsmath}
\usepackage{amssymb}
\usepackage{multirow}
\usepackage{array}
\usepackage{tabularx}
\usepackage{caption}
\usepackage{subcaption}
\usepackage{hyperref}
\hypersetup{
    colorlinks=true,
    linkcolor=blue,
    filecolor=magenta,      
    urlcolor=cyan,
    pdftitle={Overleaf Example},
    pdfpagemode=FullScreen,
    }

\newcolumntype{P}[1]{>{\centering\arraybackslash}p{#1}}

\begin{document}

\title{Enhancing 3D-Air Signature by Pen Tip Tail Trajectory Awareness: Dataset and Featuring by Novel Spatio-temporal CNN}


\author{
Saurabh Atreya, Maheswar Bora, Aritra Mukherjee, Abhijit Das\\
Birla Institute of Technology and Science, Pilani – Hyderabad Campus \\
Secunderabad, Telangana 500078\\
{\tt\small abhijit.das@hyderabad.bits-pilani.ac.in}
}

\maketitle
\thispagestyle{empty}

\begin{abstract}
 This work proposes a novel process of using pen tip and tail 3D trajectory for air signature. To acquire the trajectories we developed a new pen tool and a stereo camera was used. We proposed SliT-CNN, a novel 2D spatial-temporal convolutional neural network (CNN) for better featuring of the air signature. In addition, we also collected an air signature dataset from $45$ signers. Skilled forgery signatures per user are also collected. A detailed benchmarking of the proposed dataset using existing techniques and proposed CNN on existing and proposed dataset exhibit the effectiveness of our methodology. 
\end{abstract}


\vspace{-4mm}
\section{Introduction}

Depending on the acquiring setup, signature biometrics methods can be broadly classified into three categories \textit{offline, online, and air}. \textit{Offline} scenario is the most traditional one where the signature is done on paper using a pen and then scanned as an image. Therefore, the signature comes up as 2D data \cite{plamondon2000online, das2016multi}. Hence techniques such as Support Vector Machine (SVM), Pixel Matching (PM), Artificial Neural Network (ANN), CNN have been explored in the literature. 

In the case of \textit{an online} scenario, instead of a paper, a tablet or a pad is used as a surface to acquire a digital signature using an e-pen or fingertip movement on a digital display screen \cite{plamondon2000online}. Therefore, online signature records dynamic temporal features of the signature such as pen pressure and speed. In literature, techniques like Hidden Markov Model (HMM), SVM, and  Dynamic Time Warping (DTW) are usually explored.  

To date, traditional online and offline signature mechanisms remain widely accepted in practice. In fact, online signatures are found to be more robust and less vulnerable than their offline counterparts as the online signature has more information. Moreover, mimicking the shape of a signature is a relatively easy task in comparison to the dynamic behavioral information captured in online modes. Further, we can add a third dimension of temporal information as a hand gesture while performing the signature that can yield better advantages over existing online/offline signature biometrics. Owing to this very reason and to make signature biometrics ready for augmented reality environments, touchless forms of biometrics, air signatures were introduced \cite{behera2018analysis}.

\textit{Air signatures} are captured from the finger or hand gestures of individuals. Three forms of acquiring setup are adopted in the literature: from an egocentric viewpoint using a Google Glass headset \cite{sajid2015vsig}, from 3rd person's view by placing a camera placed in front of the subject \cite{malik20183dairsig} and using an accelerometer readings in a cell phone \cite{bailador2011analysis}. For featuring air signature  most of the previous works have used DTW \cite{malik20183dairsig, fang2017novel}. In these initial works, air signatures are recorded using an RGB sensor which lacks depth information. As 3D information is paramount in air signature researchers adopted depth cameras to capture air signatures to address this. Malik \etal  \cite{malik20183dairsig} introduced the air signature acquisition method using a single depth camera. Further, \cite{behera2018analysis} uses a leap motion sensor to capture depth information from air signature. In Zhang \etal \cite{zhang2013new}, the authors used Microsoft Kinect-based color and depth video sequences to perform recognition of 3D digits and characters. Several types of techniques such as Multi-Layer Perceptron (MLP)\cite{ketabdar2010magisign}, LSTM, HMM \cite{behera2018analysis}, Siamese network with LSTM+1D-deep CNN \cite{ghosh2021novel} are explored in the literature. Air signature biometrics is a relatively new signature modality that is introduced in recent literature. Regardless of the recent advances, the performance of air signature is not comparable to offline and online signature modalities. 

It is to be noted that air signature modality inherently contains additional depth information in the third dimension. In contrast to traditional approaches of capturing signature, the lack of a firm writing plane and visual feedback leads to large variations between signatures of the same individual. Naturally, a high amount of variation imparts additional challenges for pattern recognition. Incidentally, in previous works, only the pen tip trajectory has been engaged for air signature biometrics. Deviating from that philosophy, in this work we attempted to investigate the \textit{potential of the combination of unique 3D trajectory pattern of the pen tail and tip} (refer to Figure~\ref{fig:different_files}(c)) to underpin the highlighted challenge. We hypothesize that the combination of the 3D trajectory of the pen tail along with the tip will bring in extra information such as the orientation of the hand gesture. In addition, instead of using finger/pen tips, our solution also helps in case of occlusions. 

Another challenge that exists is the dependency on existing setup/device for air signature acquisition. Usually, an RGB or depth camera, a wearable camera such as Google Glass, or a movement sensor in a cell phone is employed. However, using these sensors is a costly process. To mitigate this challenge, we proposed a novel setup for 3D signature capturing via a stereo-vision camera. The solution is very economical as any dual camera can be used in this context and is suitable for real-time applications. In principle, we acquire the disparity in the kinematic structure of the pen head and tail to materialize the 3D signature. 

To justify the validation of our hypothesis of using pen tip and tail for air signature and the proposed acquisition setup, we developed a dataset with genuine and skilled forged signatures. In addition, we also benchmark the proposed dataset with various featuring techniques such as LSTM, 1D-CNN, and gated recurrent unit (GRU). Nevertheless, existing benchmarking techniques only concentrate on the temporal sequence of the 3D trajectory. In reality, the 3D trajectory of the air signature includes both spatial and temporal information, in two dimensions. Hence, to utilize this property, we developed a novel 2D CNN architecture (SliTCNN) that convoluted over the 2D representation of the air signature trajectory. 

\begin{figure}[h]
    \centering
    \includesvg[width=8.4cm]{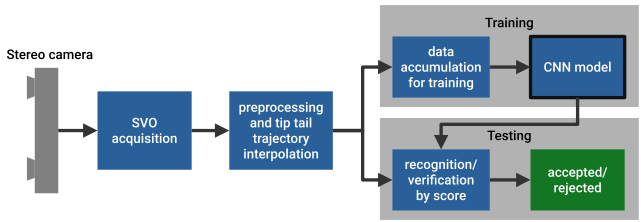}
    \caption{Overall pipeline of proposed methodology. SVO is the file format of calibrated stereo video by Zed}
    \label{fig:pipeline}
\end{figure}

To summarise we collected the data using a stereo camera using the proposed setup and the pen tool (see Figure~\ref{setup}). Further, we pre-process the data, which was used to model the air signature. The overall pipeline of the proposed approach is described in Figure~\ref{fig:pipeline}.

The specific contributions of this paper are as follows:
\begin{itemize}
    \item A novel process of using pen tip and tail depth trajectory for air signature and a stereo-camera-based capturing setup.
    \item A new spatial-temporal 2D-CNN for better featuring of the air signature. Additionally, detailed benchmarking on the proposed dataset and existing datasets using proposed and existing techniques.
    \item A novel air signature dataset that was captured using a stereo camera that has both the pen tip and tail depth dynamics.
    
\end{itemize}

\section{Proposed methodology for Air Signature}
In this section, we aim to explain the details of the proposed air signature-capturing setup, pre-processing techniques, and the novel SliTCNN architecture. 

\subsection{Proposed air signature capturing set-up}
As mentioned previously, the proposed air signature uses the 3D trajectory of both the pen tip and the pen tail as the input feature. For recording the 3D trajectory of the signatures, we have used stereo vision and a simple tool to capture the trajectory at an acceptable frame rate. We proceed to discuss the employed stereo camera, the pen tool, and the capturing protocols in detail.
\vspace{-4mm}
\subsubsection{The stereo camera}
The stereo camera used in this work is Zed2\cite{zed2} by stereo labs and the available free SDK provided by them is used for calibration and fast data capture using an embedded GPU. The mode used for the capturing is VGA with a maximum capture rate of $100$ FPS. The frame rate of the eventual capture drops to no less than $60$. The drop in frame rate is due to rendering the streams of multiple kinds of feedback from the capturing setup see Figure~\ref{setup}) provided to the user while capturing the signature. The lens distortion correction and epipolar alignment are provided by the SDK\cite{zedsdk} when calibration is done (one time). In the experiments, the auto white balance provided by the SDK is used without any external tuning. 
\vspace{-4mm}
\subsubsection{The pen tool}
In most of the previous works related to air signature, the tip of the index finger of the hand is used as the pen tip\cite{fang2017novel,behera2018analysis,malik20183dairsig,sajid2015vsig,behera2018air,zhang2013new}. We propose a novel approach to capturing both the pen tip and tail to include the writing style of the user as an added feature. For that, we propose to use a simple low-cost tool that does not consume any power. The tool is composed of a simple stick with two distinctly colored balls attached at the ends. For our experiments, we implemented the tool with a pencil and two rubber balls (of green and orange color) adhered to the ends so that they can be identified easily with some constraints on the capturing protocol. The pen-tip is of orange color and the pen-tail is of green color.

\begin{figure}[htbp]
  \centering
  \includesvg[width=7.5cm]{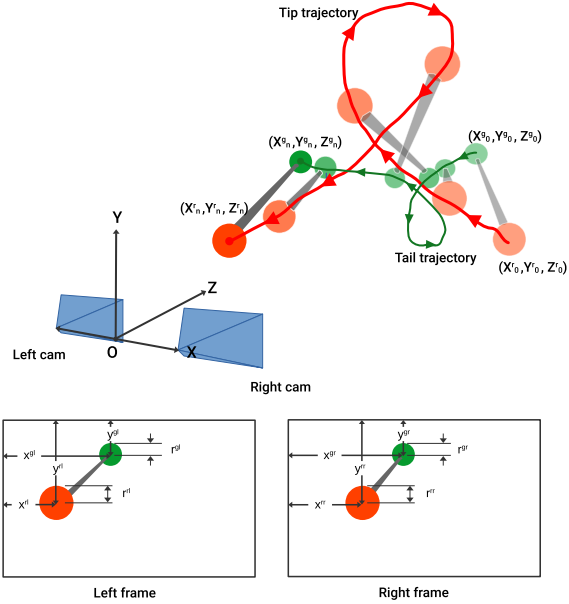}
  \caption{Schematic diagram of the signature capturing process. The frames above show the final pen position (no transparency) as seen by the stereo cameras. }
  \label{skematic_fig}
\end{figure}

\vspace{-8mm}
\subsubsection{The capturing setup and protocols}
\begin{figure}[h]
    \centering
    \begin{subfigure}[b]{0.48\columnwidth}
         \centering
         \includegraphics[width=\textwidth]{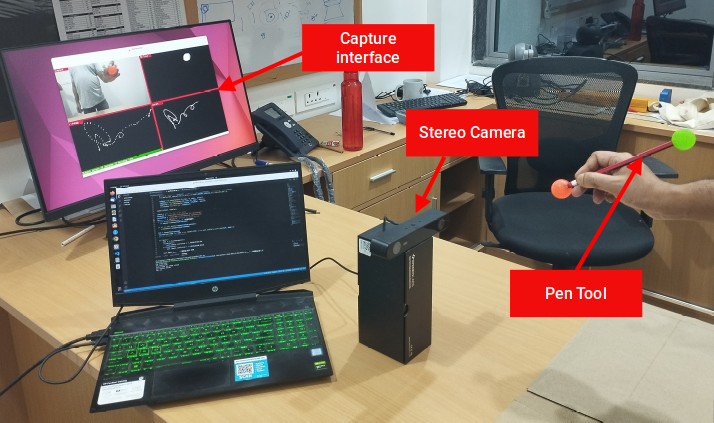}
         \caption{}
     \end{subfigure}
     \hfill
     \begin{subfigure}[b]{0.48\columnwidth}
         \centering
         \includegraphics[width=\textwidth]{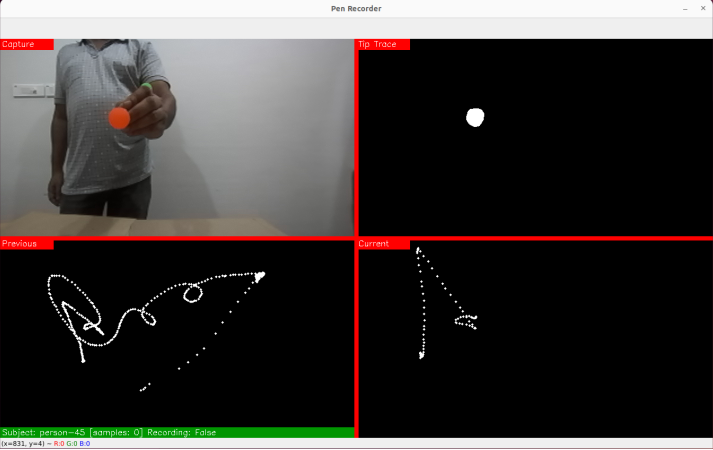}
         \caption{}
     \end{subfigure}  
    \caption{The proposed capturing setup and a screenshot of the software (a)Entire setup for air signature acquisition (b)Screenshot of capturing software}
    \label{setup}
\end{figure}

The capturing setup consists of the stereo camera along with a system having an NVIDIA GPU for high-speed processing of the stereo frames through the provided SDK (see Figure~\ref{setup}(a). A fixed protocol was maintained for collecting our data in an indoor environment. 
A recording software was developed to give the view of the left camera, the mask of the detected orange ball, the 2D trace progress of the ongoing signature and the previous signature's 2D trace (see Figure~\ref{setup}(b)). The starting and stopping time stamps for capturing the signature were regulated manually by an operator. The start and the stop time were promoted by the signer. \\

The capturing software recorded the stereo frames as SVO files so that they can be parsed using the ZED SDK without the requirement of camera calibrations. The trace of the pen tip was drawn as a series of thick dots on the center of the detected orange ball, for all the frames. 
To ensure high frame rate pre-processing and featuring were handled after recording. 

\subsection{Pre-processing of air signature}
The signatures were captured as stereo frames and then the 3D positional information was extracted from the recorded frames. To detect the balls, a simple color bandpass filter was used to detect the green and orange pixels and then morphological operations were performed to eliminate speckles. The masks for the orange and green balls for both the left and right frames were determined by fitting circles\cite{davies1988modified} on all the components and then considering the circle with the largest radius. Ultimately the signatures were saved as 3D trajectory in CSV files where each row represents a stereo-frame and there are $12$ columns that represent the 3D positional information of the two balls. Each row contains the center and radius i.e. the feature of both the balls as tuples of $(x,y,r)$ for the left and right frames. In Fig \ref{skematic_fig} the $12$ features are shown as $(x^{gl},y^{gl},r^{gl})$ for the left frame green ball, $(x^{rl},y^{rl},r^{rl})$ for the left frame orange ball, $(x^{gr},y^{gr},r^{gr})$ for the right frame green ball and $(x^{rr},y^{rr},r^{rr})$ for the right frame orange ball. Further, the features are normalized in the range of $0-1$ by dividing the $x$ with the frame-width, and the $y$ by the frame-height. The radius $r$ is normalized by dividing by the frame height. Thus CSV files for each signature have a fixed number of columns and a variable number of rows depending on the length and speed of the signer. In some timestamps the green ball or the pen tail was occluded by the hand while performing the signature, the tuple for those frames was represented as $(-1,-1,-1)$ for both the left and right frame. \\
\begin{figure}[h]
    \centering
    \begin{subfigure}[b]{0.3\columnwidth}
         \centering
         \includegraphics[width=\textwidth]{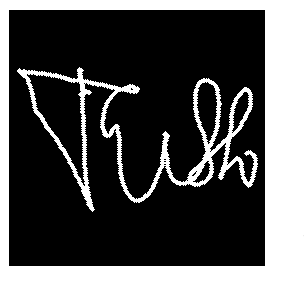}
         \caption{}
     \end{subfigure}
     \hfill
     \begin{subfigure}[b]{0.34\columnwidth}
         \centering
         \includegraphics[width=\textwidth]{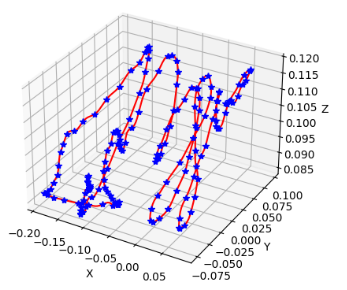}
         \caption{}
     \end{subfigure}  
     \hfill
     \begin{subfigure}[b]{0.34\columnwidth}
         \centering
         \includegraphics[width=\textwidth]{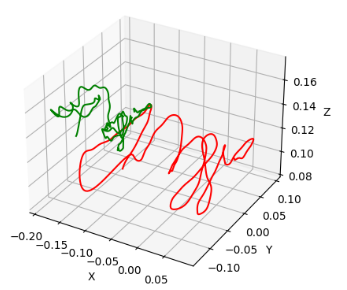}
         \caption{}
     \end{subfigure}

    \caption{The different forms of signature used in experiments: (a)interpolated 2D tip trace (b)interpolated 3D tip trajectory in red with the sampled positions in blue stars (c)interpolated 3D tip and tail trajectory with tip in red and tail in green }
    \label{fig:different_files}
\end{figure}
While parsing the stereo frames four kinds of files are generated for each signature for different kinds of experiments as depicted in Fig~\ref{fig:different_files}. They are as following:
\begin{itemize}
    \item \textit{Raw sequence:} The csv file with all $12$ columns and rows equal to the number of frames. 
    
    \item \textit{Interpolated 2D image:} The png file that contains the 2D trace of the signature from the left frame interpolated to a fixed number of points by a B-spline as defined by Diercx and Paul\cite{dierckx1995curve}(See Figure ~\ref{fig:different_files}(a)).
    
    \item \textit{Tip-tail trajectory:} The csv file with the $(X,Y,Z)$ tuple of both the balls in a column, where $X,Y,Z$ is the 3D coordinates of a ball center in the metric space. 
    Stereo disparity, (i.e. $x^{gr}-x^{gl}$ for green and $x^{rr}-x^{rl}$ for orange ) of balls are used to calculate the 3D coordinates. Frames, where the green ball is occluded, are dropped while parsing. Hence in this format number of columns is $6$ (i.e. $(X^r,Y^r,Z^r,X^g,Y^g,Z^g)$ as per Fig \ref{skematic_fig}) and the number of rows is dependent on the total number of frames discarding the number of frames in which the green ball is not detected. 
    
    \item \textit{Interpolated tip-tail trajectory:} The csv file that contains the 3D tip and tail trajectories of the signature is interpolated to a fixed number of points by a 3D B-spline curve as defined by Diercx and Paul\cite{dierckx1995curve} (See Figure ~\ref{fig:different_files}(b-c)).
\end{itemize}

\subsection{SliTCNN for Air Signature}

\begin{figure}[h]
    \centering
    \includesvg[width=5.8cm]{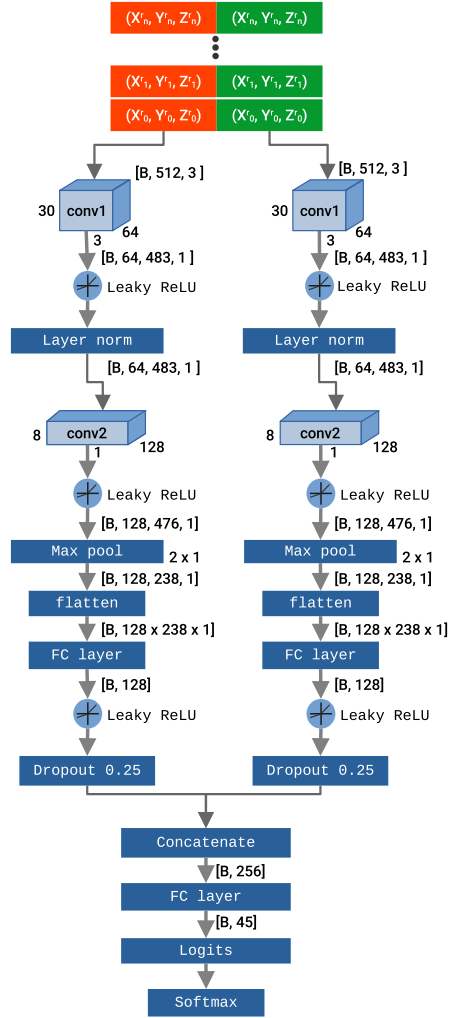}
    \caption{The detailed network architecture of the proposed 2-stream SliTCNN.}
    \label{fig:architecture}
\end{figure}

It is evident from previous literature that significant advancements have been made to model air signature. A major shortcoming that can be found is that most of the techniques considered only the temporal synergy of the data, the spatial aspect of the data is not given any leverage. Although for air signatures, spatiotemporal synergy is very important while modeling. Each air signature sample can be perceived as 2D spatiotemporal data, where each spatial data point consists of the $(x,y,z)$s of the orange and green ball, and the other dimension is along time. Inspired by \cite{ullah2021hybrid}, to manifest spatiotemporal synergy we attempted to explore the power of 2D CNN to underpin the gap. The proposed \textbf{Sli}ding \textbf{T}ime \textbf{CNN} (SliTCNN) slides 2D spatiotemporal kernels over time. To explore the information from both balls independently and then fusing their independent learned features, we took a 2-stream approach. The block diagram of the proposed 2-stream SliTCNN architecture for air signature is in Figure ~\ref{fig:architecture}.

The input to a CNN should be of a fixed dimension, whereas the temporal dimension of the signature will vary depending on the time taken by the signer for each signature. Hence, we interpolated the data to a fixed timestamp $t$ (as mentioned in )section 2.2 interpolated pen tail tip trajectory). Therefore, the input dimension to each stream of the CNN is $t \times 3$ where $3$ comes from the tuple size of $(X,Y,Z)$. While considering the signature from the proposed pan tail-tip-trajectory, two discreet signature patterns can be found (see Figure ~\ref{skematic_fig}). One of the signatures, the pattern is from the pen's tail, and the other is from the tip. Hence, we considered learning the signature from both patterns independently and then combining them. So we proposed to use a two-stream SliTCNN, where the input to one stream is the 3D points sequence of the orange ball i.e. pen tip, and in the other stream 3D points sequence of the green ball i.e. pen tail. 

For both the stream in the first two layers, we employed 2D convolution. For the first layer, $3 \times 30$ convolution with a stride of one and feature size $64$ is used, For the second layer, $1 \times 8$ with a stride of one and feature size $128$. Both convolutions are done along the temporal dimension. In turn, this convolution ensures the spatial and temporal featuring. The convolutions can be interpreted in a way that it is analyzing $n$ timestamps at a time while convolving ($30$ temporal points for 1st layer), hence it can manifest both the temporal and spatial feature it witness in that course. 

The convolution layers are followed by the activation layer. In the first convolution layer normalization is added after activation. Whereas, the second convolution is max pooling by $2\times 1$ after activation. The max pooled output of the second convolution layers from both the streams are being flattened followed by activation with Leaky ReLU and then concatenated. The concatenated feature is passed to a linear layer of size $512$ which is followed by the output layer with Softmax activation to get the final prediction.  

We considered Leaky ReLU as the activation as among the data $X,Y$s can be negative. Hence the feature maps can have negative values and restoring them with Leaky ReLU activation will help the gradient flow and thereby the optimization. Similarly, layer normalization is selected as normalizing each of the features in the batch independently across all features will make sense in our context as that will restore the spatiotemporal synergy. Before flattening max pooling is employed because it is necessary to aggregate global features before fusing the feature from both streams.

The output from the two streams passes through individual fully connected (FC) layers before being concatenated. This enables a stream-wise latent representation which is fed into a final FC layer before being passed through Softmax to get the class score. A dropout with $0.25$ is applied before the concatenation which prevents the final FC layer from overfitting.

\section{Experimental results}

In this section, we introduce the proposed dataset, describe the other datasets used, implementation details, and benchmarking techniques, and at last a critical discussion of the experimental results. 

\subsection{Detailed Description of the Proposed Datasets}

\begin{figure}[h]
    \centering
    \begin{subfigure}[b]{0.48\columnwidth}
         \centering
         \includegraphics[width=\textwidth]{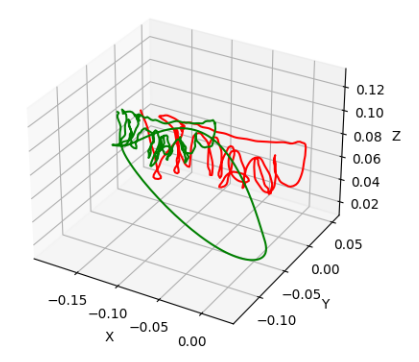}
         \caption{}
     \end{subfigure}
     \hfill
     \begin{subfigure}[b]{0.48\columnwidth}
         \centering
         \includegraphics[width=\textwidth]{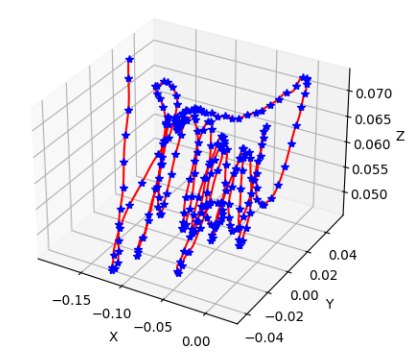}
         \caption{}
     \end{subfigure}  
     \begin{subfigure}[b]{0.48\columnwidth}
         \centering
         \includegraphics[width=\textwidth]{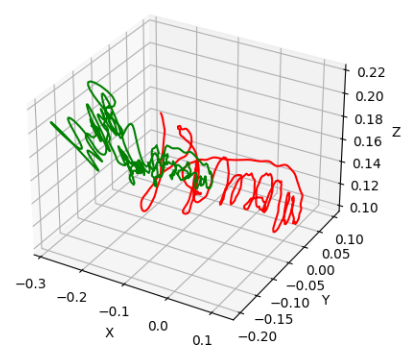}
         \caption{}
     \end{subfigure}
     \hfill
     \begin{subfigure}[b]{0.48\columnwidth}
         \centering
         \includegraphics[width=\textwidth]{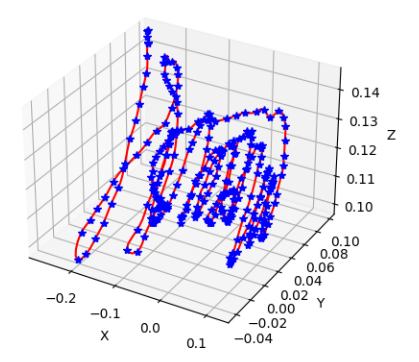}
         \caption{}
     \end{subfigure} 

    \caption{Why tip-tail is good: (a) genuine signature of a participant in tip-tail format, (b) the pen-tip format for the same signature (c) skilled forgery of the same participant in tip-tail format and (d) the pen-tip format for the same forged signature. Note that it is comparatively easy to mimic the pen-tip trajectory in air signature (similarity of b and d) but to do so for the tip-tail is difficult (see dissimilarity between a and c) }
    \label{fig:genuine_vs_forged}
\end{figure}

The proposed air signature dataset T3AAS-v1 was recorded at the author's institute following all necessary ethical rules and permissions. English signatures from $45$ volunteers were collected via the given setup in Section 2.1. For each signer, $25$ signatures were collected in one session. Corresponding $12$ skilled forgeries per signer were also developed. Each forger was allowed to practice the signature for as long as they were required to produce the forgery. Each forger imitated the $12$ signatures of $5$ signers each day. The genuine signatures shown to each forger are chosen randomly from the $25$ genuine ones. For each signer among $25$ genuine signatures, $16$ was used for training, $4$ for validation, and the rest for testing. An example of genuine and its corresponding forgery signature represented in 3D spline is in Figure~\ref{fig:genuine_vs_forged}.

\subsection{Other Datasets used for Comparison}

To compare the robustness of our proposed methodology we also experimented with one air signature and one online signature dataset available in the literature.

\textbf{LeapMotionAirSignature} \cite{behera2018analysis}: The dataset has $14$ air signatures per signer from $50$ users. $8$ samples were used for training, $2$ used for validation, and the rest for testing. 

\textbf{SVC 2004} \cite{yeung2004svc2004} is an online database containing the Chinese and Roman script signatures of $40$ users. Among $20$ samples from each user $10$ was used for training, $2$ was used for validation, and the rest $8$ for testing. The $20$ skilled forgeries were also utilized for forgery experiments. 

\begin{table*}[ht!]
\footnotesize
\caption{Details of benchmarking techniques input form and a brief description of the techniques}
\label{benchtech1}
\begin{tabular}{|p{4.3cm}|p{2cm}|p{9.9cm}|}
\hline
\textbf{Methodology}            & \textbf{Input}        & \textbf{Description}                                                                  \\ \hline
Interpolated TipTail SimpleGRU &
  \multirow{10}{*}{\parbox{2cm}{Interpolated pen tip or tiptail trajectory}} &
  \multirow{2}{*}{Data fed together into a single vanilla GRU with FC layer to output} \\ \cline{1-1}
Interpolated PenTip SimpleGRU   &                       &                                                                                       \\ \cline{1-1} \cline{3-3} 
Interpolated TipTail 2 level GRU &
   &
  Data fed tip and tail wise in two GRU units and result is concatenated and sent through a GRU unit before FC layer to output \\ \cline{1-1} \cline{3-3} 
Interpolated TipTailSimpleLSTM  &                       & \multirow{2}{*}{Data fed together into a single vanilla LSTM with FC layer to output} \\ \cline{1-1}
Interpolated PenTip SimpleLSTM  &                       &                                                                                       \\ \cline{1-1} \cline{3-3} 
Interpolated TipTail 2 level LSTM &
   &
  Data fed tip and tail wise in two LSTM units and the result is concatenated and sent through an LSTM unit before FC layer to output \\ \cline{1-1} \cline{3-3} 
Interpolated TipTail CNN1D &
   &
  \multirow{2}{*}{Data fed coordinate-wise 1D temporal convolution kernels} \\ \cline{1-1}
Interpolated PenTip CNN1D       &                       &                                                                                       \\ \cline{1-1} \cline{3-3} 
Proposed 2 stream SlitCNN       &                       & Data into 2 streams of the CNN spatiotemporally for each ball (see Figure~\ref{fig:architecture})                              \\ \cline{1-1} \cline{3-3} 
SliTCNN                         &                       & Data of both tip and tail fed together in a single stream instead, spatiotemporal kernel covers the entire span of features (1st conv layer filter size =$6\times 30$, 2nd conv layer filter size = $1\times8$,)                           \\ \hline
Interpolated Pentip trace VGG16 & Interpolated 2D image & Data fed to standard VGGnet 16 with last FC layer tuned for classes                   \\ \hline
\end{tabular}%
\end{table*}
\vspace{-5.1mm}

\subsection{Implementation details}
For all experiments, Adam optimizer with $\beta 1 =0.9$, $\beta 2=0.999$, and a learning rate of $0.00001$ was used. For all models, the cross-entropy loss function is used. The model performing best on the validation set is chosen for testing. The implementation and the dataset are available at: \href{https://github.com/atrey-a/T3AAS-v1-benchmarking}{github.com/atrey-a/T3AAS-v1-benchmarking}.

For respective experiments, the data were interpolated to a spline length of $512$ using the interpolated method explained in Section 2.2. The original data is rotated in turns by the following angles (in degrees): $-10,-5,0,5,10$. Following that, they are scaled in turns  along $x$ and $z$, or $y$ and $z$ by factors of $0.95, 1, 1.05$. This gives us a total of $30$ times the original data after augmentation.

\subsection{Benchmarking techniques}
The proposed dataset is benchmarked with various featuring techniques used in the literature such as LSTM, 1D-CNN, GRU and the proposed SliTCNN. The lists of benchmarking techniques along with their various variations employed, input to these techniques, and their brief description are in Table~\ref{benchtech1}.

\subsection{Results and discussion}
Now we proceed to discuss the benchmarking results and the detailed ablation study on the proposed SliTCNN.  
\vspace{-2mm}
\subsubsection{Benchmarking results}
The results of different aforementioned benchmarking techniques on T3AAS-v1,  LeapMotionAirSignature, and SVC 2004 are in Table~\ref{benchmark_ours}, ~\ref{benchmark_svc} and ~\ref{benchmark_leap} respectively. For all the results reported the proposed augmentation technique was used. It can be observed from all the tables that the proposed SliTCNN has produced the best results for all the datasets except VGG16, considering both recognition and verification tasks (except for the random forgery experiment of LeapMotionAirSignature Simple GRU worked best). Though VGG16 gave better results but it should be noted that the parameter size of VGG16 is more than $16$ times that of the proposed SliTCNN. Further, for T3AAS-v1 2-stream version of SliTCNN has performed better than SliTCNN. 

\begin{figure*}[hbt!]
    \centering
    \begin{subfigure}[b]{0.33\textwidth}
         \centering
         \includegraphics[width=\textwidth]{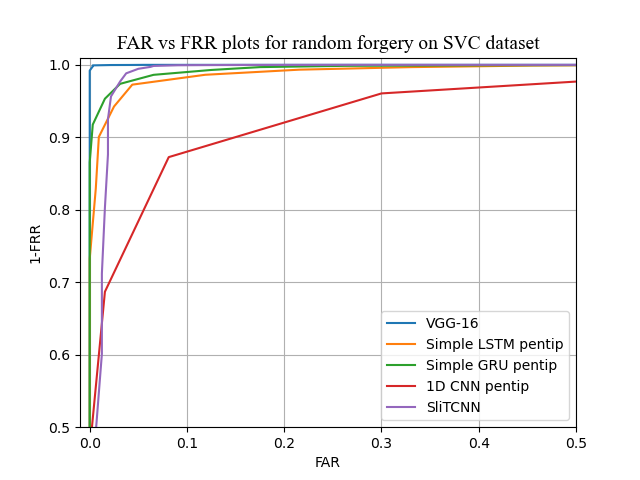}
         \caption{}
     \end{subfigure}
     \hfill
     \begin{subfigure}[b]{0.33\textwidth}
         \centering
         \includegraphics[width=\textwidth]{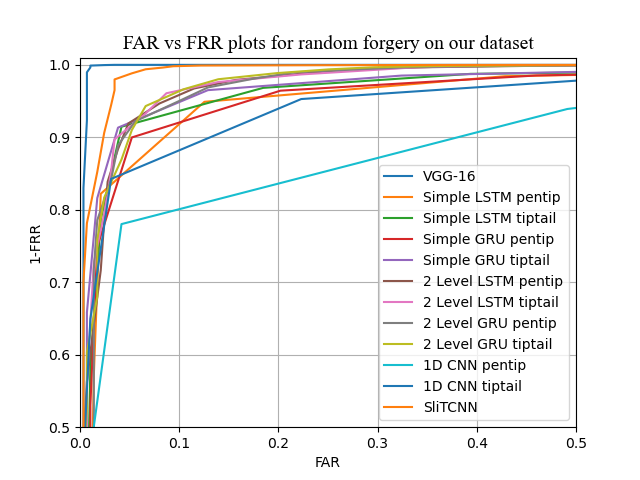}
         \caption{}
     \end{subfigure}  
     \hfill
     \begin{subfigure}[b]{0.33\textwidth}
         \centering
         \includegraphics[width=\textwidth]{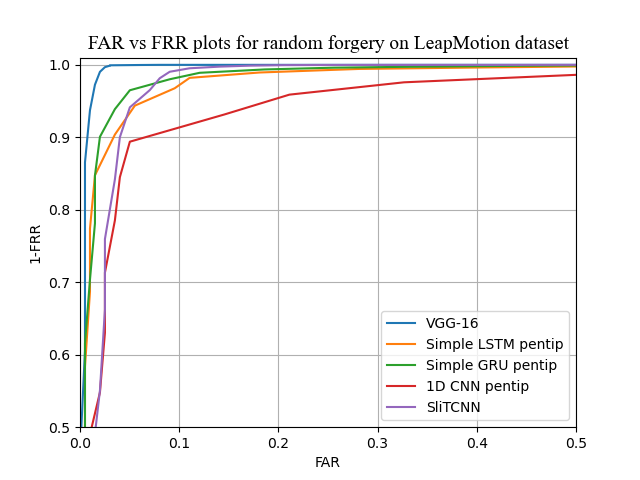}
         \caption{}
     \end{subfigure}
     
     \begin{subfigure}[b]{0.33\textwidth}
         \centering
         \includegraphics[width=\textwidth]{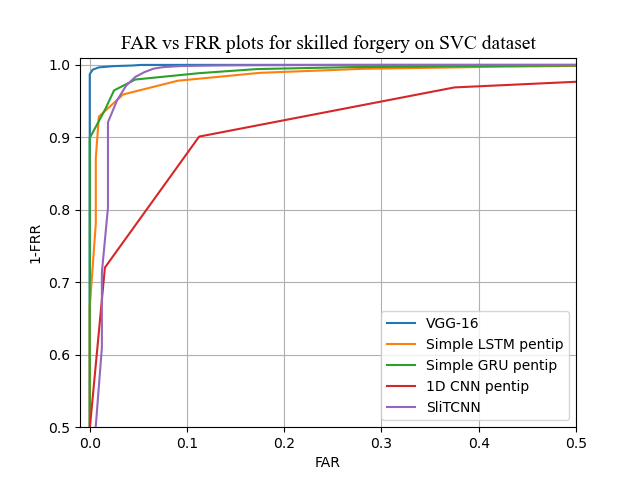}
         \caption{}
     \end{subfigure} 
     \hfill
     \begin{subfigure}[b]{0.33\textwidth}
         \centering
         \includegraphics[width=\textwidth]{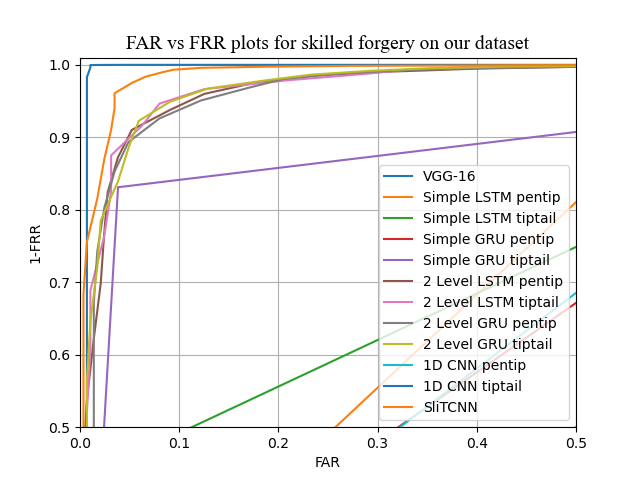}
         \caption{}
     \end{subfigure}
     \hfill
     \begin{subfigure}[b]{0.33\textwidth}
         \centering
         \includegraphics[width=\textwidth]{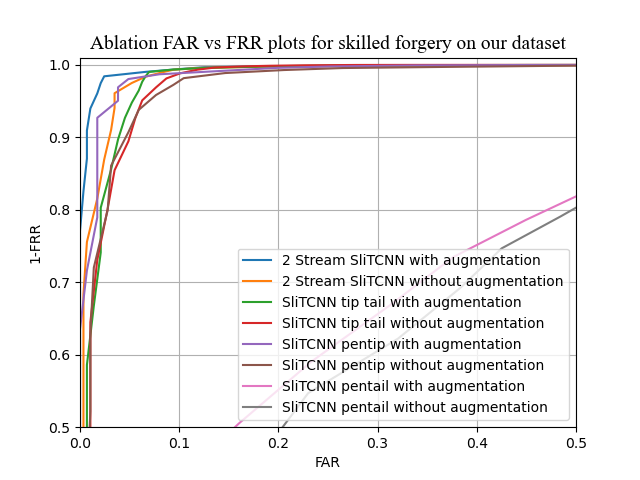}
         \caption{}
     \end{subfigure}

    \caption{ROC curves for different methodologies on all the datasets }
    \label{roc}
\end{figure*}

\begin{table*}[h]
\small
\caption{Bechmarking results over proposed T3AAS-v1 dataset on different methodologies with augmentation}
\label{benchmark_ours}
\renewcommand{\arraystretch}{1.3}
\resizebox{\textwidth}{!}{%
\begin{tabular}{|P{0.25\textwidth}|P{0.15\textwidth}|P{0.15\textwidth}|P{0.15\textwidth}|P{0.15\textwidth}|P{0.15\textwidth}|}
\hline
 & Recognition accuracy in \% & Parameter size (in millions) & Testing time (in microsecs) & EER Random Forgeries in \% & EER Skilled Forgeries in \% \\ \hline
Interpolated TipTail SimpleGRU  & 90.94 & 4.67   & 9083.95   & 6.25  & 10.36 \\ \hline
Interpolated PenTip SimpleGRU   & 86.76 & 4.67   & 12139.03  & 7.62  & 40.4  \\ \hline

Interpolated TipTail 2 level GRU &	83.27 & 14.5&	2932.36	& 6.19	& 7.33 \\ \hline
Interpolated Pentip 2 level GRU &	80.84	& 8.1&	2379.53	& 6.41	& 7.7 \\ \hline
Interpolated TipTailSimpleLSTM  & 89.2  & 5.2    & 15318.02  & 6.7   & 30.87 \\ \hline
Interpolated PenTip SimpleLSTM  & 85.71 & 5.2    & 15200.69  & 7.27  & 31.87 \\ \hline
Interpolated TipTail 2 level LSTM	& 77.7	& 16.2	&3251.26	& 8.7	& 8.78 \\ \hline
Interpolated PenTip 2 level LSTM	& 76.3	& 16.2	&3279.16	& 8.84 & 9.56 \\ \hline
Interpolated TipTail CNN1D      & 83.97 & 0.19   & 1308.54   & 9.44  & 50.05 \\ \hline
Interpolated PenTip CNN1D       & 76.65 & 0.12   & 1225.84   & 13.08 & 40.15 \\ \hline
SliTCNN & 93.72 &  4.04 & 2077.21 &  4.68 & 5.21\\\hline
Proposed 2 stream SliTCNN       & \textbf{97.21} & 8.01   & 2151.73   & \textbf{1.74}  & \textbf{2.3}   \\ \hline
Interpolated Pentip trace VGG16 & 98.6  & 134.44 & 128329.73 & 0.86  & 0.71  \\ \hline
\end{tabular}%
}
\end{table*}

\begin{table*}[h]
\small
\caption{Bechmarking results over SVC dataset~\cite{yeung2004svc2004} on different methodologies with augmentation for only pen tip in 2D}
\label{benchmark_svc}
\renewcommand{\arraystretch}{1.3}
\resizebox{\textwidth}{!}{%
\begin{tabular}{|P{0.25\textwidth}|P{0.15\textwidth}|P{0.15\textwidth}|P{0.15\textwidth}|P{0.15\textwidth}|P{0.15\textwidth}|}
\hline
 & Recognition accuracy in \% & Parameter size (in millions) & Testing time (in micro seconds) & EER Random Forgeries in \% & EER Skilled Forgeries in \% \\ \hline
Interpolated PenTip SimpleGRU   & 96.87 & 6.9    & 18331.02 & 2.87 & 3.013 \\ \hline
Interpolated PenTip SimpleLSTM  & 96.25 & 5.2    & 34191.26 & 3.55   & 3.76   \\ \hline
Interpolated PenTip CNN1D       & 90.6  & 0.12   & 1698.25  & 10.43  & 10.57  \\ \hline
Single stream SliTCNN       & \textbf{95.94} & 8.05   & 3909.18  & \textbf{2.71 }  & \textbf{3.3}    \\ \hline
Interpolated Pentip trace VGG16 & 99.37 & 134.44 & 112157.7 & 0.26 & 0.49 \\ \hline
\end{tabular}%

}
\end{table*}

\begin{table*}[ht!]
\small
\caption{Bechmarking results over LeapMotionAirSignature dataset~\cite{behera2018analysis} on different methodologies with augmentation for only pen tip in 3D and 2D but without skilled forgery}
\label{benchmark_leap}
\renewcommand{\arraystretch}{1.3}
\resizebox{\textwidth}{!}{%
\begin{tabular}{|P{0.4\textwidth}|P{0.12\textwidth}|P{0.12\textwidth}|P{0.15\textwidth}|P{0.15\textwidth}|}
\hline
 & Recognition accuracy in \% & Parameter size (in millions) & Testing time (in microsecs) & EER Random Forgeries in \% \\ \hline
Interpolated PenTip SimpleGRU   & 89.95 & 14.8   & 40388.45  & \textbf{4.27 }  \\ \hline
Interpolated PenTip SimpleLSTM  & 89.95 & 15.3   & 60648.37  & 4.65 \\ \hline
Interpolated PenTip CNN1D       & 76.39 & 0.31   & 2555.19   & 7.81 \\ \hline
Single stream SliTCNN       & \textbf{90.95} & 8.01   & 6819.56       &  5.44   \\ \hline
Interpolated Pentip trace VGG16 & 97.49 & 134.46 & 140802.38 & 1.49   \\ \hline
\end{tabular}%
}
\end{table*}

\begin{table*}[ht!]
\small
\centering
\caption{Detailed ablation study of proposed SliTCNN}
\label{ablation}

\renewcommand{\arraystretch}{1}
\resizebox{\textwidth}{!}{%

\begin{tabular}
{|P{0.39\linewidth}|P{0.1\linewidth}|P{0.1\linewidth}|P{0.1\linewidth}|P{0.09\linewidth}|P{0.09\linewidth}|}

\hline
Models &
  Parameter size in millions &
  Testing time (in micro secs) &
  Recognition accuracy in \% &
  EER Random Forgeries in \% &
  EER Skilled Forgeries in \% \\ \hline
SliTCNN with pen tail (without augmentation)      & \multirow{4}{*}{}4.03 & 1781.58 & 92.33 & 6.09 & 6.16 \\ \cline{1-1} \cline{4-6} 
SliTCNN with pen tail (with augmentation)         &                       &                          & 91.64 & 5.05 & 6.15 \\ \cline{1-1} \cline{4-6} 
SliTCNN with pen tip (without augmentation)       &                       &                          & 91.98 & 4.52 & 6.06 \\ \cline{1-1} \cline{4-6} 
SliTCNN with pen tip (with augmentation)          &                       &                          & 94.42 & 3.48 & 3.46 \\ \hline
SliTCNN with pen tip and tail(without augmentation) &
  \multirow{2}{*}{4.04} &
  \multirow{2}{*}{2077.21} &
  91.63 &
  5.84 &
  5.58 \\ \cline{1-1} \cline{4-6} 
SliTCNN with pen tip and tail (with augmentation) &                       &                          & 93.72 & 4.68 & 5.21 \\ \hline
Proposed 2 stream SliTCNN (without augmentation)  & \multirow{2}{*}{8.01} & \multirow{2}{*}{2151.73} & 95.47 & 3.48 & 3.7  \\ \cline{1-1} \cline{4-6} 
Proposed 2 stream SlitCNN (with augmentation)     &                       &                          & \textbf{97.21} & \textbf{1.74} & \textbf{2.3}  \\ \hline
\end{tabular}
} 
\end{table*}

While developing the SliTCNN architecture we tried several approaches such as changing the kernel size both in the spatial and temporal dimension, deploying different activation functions (ReLU, sigmoid, tanh. Leaky ReLU), different types of normalization (batch normalization and layer normalization), different pooling (max, average and min), adding more FC, increasing and decreasing the percentage of dropout nodes, completely removing all
dropout layers, adding more dropout layers at every block of FC layers, streamed version of the network, etc. The best-optimized combination in terms of performance and execution time was at last reported (see Figure~\ref{fig:architecture}).

We started the benchmarking with the existing techniques such as LSTM, GRU, 1D-CNN. VGG-16 for featuring the 2D splines of the 3D signatures was also carried out. One important fact that we observed is that LSTM, GRU, and 1D-CNN consider only the temporal synergy of the data, whereas VGG-16 considered only the spatial aspects. In fact, VGG-16 obtained better results in every aspect, hence discarding spatial information while considering temporal information will not be a good idea. Based on this observation we designed the SliTCNN. 

For LSTM, GRU, and 1D-CNN we also experimented with the raw sequence of the air signature explained in Section 2.2. For the raw sequence, the timestamp is not fixed hence end zero padding was used for GRU and LSTM, and center padding was used for 1D-CNN. The performance of the interpolated data was much better so we do not report the results with zero padding and continued to experiment with the interpolated data. One obvious reason for this betterment is owing to depth estimation being done while interpolation. Another possible reason is the fixed length of the signatures helps the temporal classifiers to learn better. 

We continued to experiment with different versions of LSTM and GRU. We employed 2-level LSTM and 2-Level GRU \cite{das2021spatio}. For 2-level LSTM and GRU, pen tail and tip are considered. In the 1st level, the interpolated data from both pen tip and tail are input to separate GRU/LSTM. The learned feature from 1st level is a fully connected layer of size 512. These features from both balls are input to the 2nd level of the GRU/LSTM. These experiments were carried out to learn the features of the pen tip and tail independently and then to fuse them but from the results, it can be concluded that these approaches were not effective. To further explore this angle, we came up with the 2-stream approach of SliTCNN. In addition to performance, owing to the light network size and low testing time of the proposed network, SliTCNN will be very handy for employment in real-life edge computing scenarios. The ROC curves of the important experiments are in Figure~\ref{roc}.

From the above discussion, we can conclude that SliTCNN has outperformed all the benchmarking techniques for all experiments and overall datasets. Also, the network size is quite comparable. The 2D pen trace approach using VGG-16 has outperformed the performance of the SliTCNN marginally but the network size is quite large. Hence, still there is room for improving the air signature models. 

\vspace{-4mm}

\subsubsection{Ablation study}
The influence of each component of our strategy is discussed through ablation experiments in Table~\ref{ablation}. To find the effectiveness of the combination of pen tail and tip we performed two sets of experiments. Once both the data $(x,y,z)$ are together fed as input to SliTCNN and once by passing the data separately in a two-stream fashion. It can be observed from Table~\ref{ablation} that by using two streams SliTCNN, the recognition accuracy, random forgery, and skilled forgery verification EER are better ~$3.5\%$, ~$4\%$, and ~$4\%$ respectively. It should also be noted that with this betterment the network size grew only by $4$M and also in terms of testing a minute increase can be observed. We also analyzed the effectiveness of the pen tip and tail separately. One interesting fact that can be noticed is that the performance of the pen tip in terms of both verification and recognition was better while using augmentation and the pen tail was better while augmentation was not used. It can also be observed that overall the augmentation technique worked well for all the reported experiments.


\section{Conclusions}
In this work, we adhere to exploring the capability of the 3D pen tip and tail trajectory for air signature biometrics. We proposed a whole set of capturing air signatures with pen tip and tail. In this regards a pen tool is developed, which is been used to sign on air in front of the stereo camera in the proposed capturing setup. To validate the effectiveness of our methodology we developed a dataset T3AAS-v1, which consists of air signatures from $45$ volunteers, along with skilled forgeries for these signatures. 

We benchmark the proposed dataset with several existing techniques. Exhaustive experimentation was performed which supports our hypothesis of using both pen tip and tail for air signature. However, it can also be concluded that none of these techniques were effective to utilize the best out of both the pen tail and tip trajectory. Hence, we proposed SliTCNN, a new 2D spatio-temporal CNN. From the ablation study, it can be observed that SliTCNN can utilize the pen tail and tip trajectory in a very effective way and its performance has outperformed any existing technique. Moreover, its performance is quite comparable to the corresponding 2D signature spline. Additionally, the network size is also quite light.

To find the effectiveness of SliTCNN and capturing methodology of our proposed air signature we made a detailed comparison with publicly available datasets. The results achieved also showcase the effectiveness of our methodology. 

\section*{Acknowledgements}

The authors gratefully acknowledge the computing time provided on the high performance
computing facility, Sharanga, at the Birla Institute of Technology and Science - Pilani, Hyderabad
Campus.

{\small
\bibliographystyle{ieee}
\bibliography{egbib}
}

\end{document}